\documentclass{article}

\usepackage{PRIMEarxiv}

\usepackage[utf8]{inputenc} 
\usepackage[T1]{fontenc}    
\usepackage[breaklinks]{hyperref}
\usepackage{url}            
\usepackage{booktabs}       
\usepackage{amsfonts}       
\usepackage{nicefrac}       
\usepackage{microtype}      
\usepackage{fancyhdr}       
\usepackage{graphicx}       
\graphicspath{{media/}}     
\usepackage{natbib}
\usepackage{multirow}
\usepackage{amsmath} 
\usepackage[USenglish]{datetime2}

\title{Reproduction and Replication of an Adversarial Stylometry Experiment
}

\author{
  Haining Wang \\
   Indiana University School of Medicine \\
   Indianapolis, Indiana\\
   \texttt{hw56@iu.edu} \\
   \And
   Patrick Juola \\
   Duquesne University \\
   Pittsburgh, Pennsylvania\\
   \texttt{juola@mathcs.duq.edu} \\
   \And
     Allen Riddell \\
   Indiana University Bloomington \\
   Bloomington, Indiana\\
   \texttt{riddella@iu.edu} \\
}

\begin{document}
\maketitle


\begin{abstract}
{Maintaining anonymity in natural language communication remains a challenging task. Even when the number of candidate authors is large, standard authorship attribution techniques that analyze writing style predict the original author with uncomfortably high accuracy. Adversarial stylometry provides a defense against authorship attribution, helping users avoid unwanted deanonymization. This paper reproduces and replicates experiments from a seminal study of defenses against authorship attribution \citep{brennan2012adversarial}. After reproducing the experiment using the original data, we then replicate the experiment by repeating the online field experiment using the procedures described in the original paper. Although we reach the same conclusion as the original paper, our results suggest that the defenses studied may be overstated in their effectiveness. This is largely due to the absence of a control group in the original study. In our replication, we find evidence suggesting that an entirely automatic method, round-trip translation, warrants re-examination because it appears to reduce the effectiveness of established authorship attribution methods.
}
\end{abstract}

\keywords{adversarial stylometry \and authorship attribution \and authorship identification \and stylometry \and writing style \and whistleblower protection}

Commercial and government entities are known to monitor the online activities of individuals through techniques such as Internet Protocol (IP) traffic analysis and behavioral fingerprinting, posing well-documented threats to user privacy.
Anonymous communication provides a counterweight to these threats by enabling individuals to communicate online without disclosing their identity. 
Through techniques such as encryption and anonymous routing, individuals can shield their online activities from being monitored and tracked.
Nevertheless, maintaining anonymity remains a challenge, even with a secure channel. This is because the analysis of an individual's writing style typically yields useful clues about their identity.

\textit{Authorship attribution} techniques allow an adversary to guess the author of an anonymous text by comparing the writing style in the unsigned text with the style found in writing samples from likely authors.
Past research shows that authorship attribution techniques tend to identify the author of an unsigned text at a rate far better than chance \citep{juola2008authorship, koppel2009computational, stamatatos2009survey}.
To achieve this rate of success, only a modest amount of pre-existing writing needs to be collected from candidate authors \citep{rao2000can, eder2015does}.
For example, previous research reports that standard authorship attribution methods achieve over 90\% accuracy given a set of 50 candidate authors \citep{abbasi2008writeprints} and 25\% accuracy given 100,000 candidates \citep{narayanan2012feasibility}.
The models used in the studies are familiar (e.g., support vector machines and naïve Bayes classifiers) and use a few hundred linguistic features extracted from a few thousand words of pre-existing writing.
For individuals who wish to remain anonymous while sharing even modest amounts of prose, standard authorship attribution methods present a serious obstacle.

Authorship attribution's threat to user privacy has arguably increased in recent years because collecting pre-existing writing samples has become easier: in an era of text-based social media, writing samples are frequently available online. In a survey conducted by the Pew Research Center, 38\% of 792 American adults reported that ``things [they] have written using [their] name[s]'' were available online \citep{rainie2013anonymity}.
It is therefore a matter of some urgency to develop and refine techniques to defeat or frustrate stylometric fingerprinting.

The defense against authorship fingerprinting, termed \textit{adversarial stylometry}, aims to prevent involuntary deanonymization by ``apply[ing] deception to writing style to affect the outcome of stylometric analysis'' \citep{brennan2012adversarial}.
Past research on adversarial stylometry has focused on manual interventions, wagering that a motivated individual may be able to alter their prose style in a new document to such a degree that standard authorship attribution methods will struggle to associate the document with the individual's previous writing.
Manual interventions to obscure one's style deserve attention even if they ultimately prove less effective than machine-supported style obfuscation methods, since they can be used when no trusted computational resources are available. Such a setting is easy to imagine in the case of whistleblowing. For example, an employee at a financial firm who seeks to expose wrongdoing while maintaining anonymity may lack immediate access to an unmonitored computer.

\paragraph{Contribution}
In this paper, we report on a reproduction and, separately, a replication of the experiment described in \citet{brennan2012adversarial}. First, we carefully \emph{reproduce} the experiment using the materials and methods described in the original paper. We perform this reproduction in order to double-check the original findings. We are motivated here by the recommendation, found in numerous academic communities, that reproduction of previous studies be regarded as an essential practice \citep{stodden2014implementing}.
Second, we \emph{replicate} the experiment using a new population of writers. In our replication, we correct for oversights in the original design, chief among them the absence of a control group.
The contributions of this work are as follows:

\begin{itemize}
    \item We confirm the usefulness of two manual adversarial strategies. Both techniques reduce the accuracy of standard authorship attribution models to around 20\% accuracy given ten candidate authors. 
    \item Our study confirms that an automatic intervention, round-trip machine translation, appears to be useful in obscuring one's writing style, though less effective than the two manual interventions.
    \item We contribute a new corpus, assembled to perform the replication, that can be used in adversarial stylometry research: the Riddell-Juola corpus.
\end{itemize}

\section{Related Works}

Following \citet{brennan2012adversarial}, we refer to techniques that aim to frustrate authorship attribution as adversarial stylometry. 
The goal of adversarial stylometry is to hide distinguishing elements of one's writing style while still communicating successfully. Ideally, the use of style obfuscation techniques should not leave conspicuous traces \citep{potthast2016author}.

In their seminal study, \citet{brennan2012adversarial} examine three strategies for obscuring an author's writing style.
Two of these strategies are manual: \textit{obfuscation}, in which an author simply writes differently than they ordinarily would; and \textit{imitation}, in which an author emulates the idiosyncratic style of a different author. The third, automated strategy is known as \textit{round-trip translation} (or ``back translation''). This strategy takes advantage of machine translation software and translates the original prose to one or more intermediate languages before returning to the original language.
The effectiveness of these three strategies was evaluated using a corpus derived from a field experiment (described in Section~\ref{sec:ebg}).
\citet{brennan2012adversarial} reported that the two manual interventions caused the performance of their most effective authorship attribution model to approach or fall below chance level. 
Although round-trip translation was less effective, it still meaningfully hindered authorship attribution.

Since the publication of the study, numerous automatic approaches have been proposed. Round-trip translation, in particular, has been widely tested with various intermediate languages \citep{brennan2012adversarial, mack2015best, day2016towards}.
In general, round-trip translation modifies an author's writing style to some degree and only slightly distorts the semantic content of the prose.
Although studies show that translation software, intermediate languages, and round-trip iteration count can be guessed \citep{caliskan2012translate, day2016towards}, this knowledge poses no immediate threat to author identity.
Notably, most studies using machine translation, including that of \citet{brennan2012adversarial}, use statistical machine translation.
More recent research using neural machine translation demonstrates that round-trip translation can effectively alter style in a way that makes guessing demographic characteristics of writers more difficult \citep{xu2019privacy, adelani2021preventing}.

Instead of using generic machine translation, attempts have been made to obscure an author's style by ``transferring'' a foreign source style to the document. For instance,  \citet{emmery2018style} employed an LSTM-based encoder-decoder translation model trained on verse pairs from distinct English versions of the Bible, including both the Old and New Testaments \citep{carlson2018evaluating}.
This model allows the translation of verses into specific styles derived from other Bible versions. 
This approach managed to generate verses that deceived a strong adversary into performing below chance levels while preserving the semantic content of the original.
When high-quality parallel data is not available, which is often the case, researchers have employed training methods that do not depend on aligned corpora.
These methods include autoencoders \citep{bakhteev2017author, bo2020authorship, weggenmann2022dp_vae}, generative adversarial networks \citep{shetty2018a4nt}, and variational autoencoders \citep{weggenmann2022dp_vae}, possibly combined with multiple \emph{ad hoc} decoders \citep{shetty2018a4nt} and disentangled representations \citep{emmery2018style}.

However, software maintenance has emerged as a significant hurdle to the widespread adoption of these automatic techniques.
For example, Anonymouth \citep{mcdonald2012use} is software that can analyze a user's writing and then offer words that should be used at a higher or lower rate in order to frustrate authorship attribution.
Despite Anonymouth being open-source and available for download, its lack of maintenance renders it virtually unusable on contemporary operating systems. 
At present, no actively maintained automatic or computer-assisted writing obfuscation tools appear to be accessible to non-technical users.

Manual circumvention, by contrast, attempts to inject unpredictable variation into an individual's writing style by prompting the author to consciously make an attempt to disguise their writing, either via writing differently or by mimicking a pre-existing style. 
Despite the impressive performance reported by \citet{brennan2012adversarial}, manual approaches remain less studied than their automated counterparts.
In addition to their independence from computational resources, another merit of manual approaches is that an individual author's tendency to write prose with author-identifying stylistic fingerprints appears to be variable, suggesting that there is considerable ``room'' for an intervention to work. Changes in topic can alter author-identifying fingerprints \citep{sapkota2014cross, altakrori2021topic}, as can changes in document genre \citep{kestemont2012cross, overdorf2016blogs}. Even the way a writer inputs a text into the computer can make a difference, e.g., via a browser's text box or via a word processor running locally \citep{wang2021mode}.
Notably, when \citet{almishari2014fighting} asked individuals from Amazon Mechanical Turk (MTurk) to rewrite a given text, they found the user-generated adversarial samples to be better than round-trip samples in terms of semantic preservation and circumventing fingerprinting. In short, untrained writers seem to have the capacity to modify texts in ways that are relevant to authorship attribution.

\section{Research Question}
We investigate how well standard authorship attribution models perform when a defensive strategy is used.
In this paper, we reproduce the study of \citet{brennan2012adversarial} using the original corpus (the Extended Brennan-Greenstadt corpus). The reproduction is described in Section~\ref{sec:reproduction}. We also replicate the experiment using a new corpus, the Riddell-Juola corpus. Section~\ref{sec:replication} describes the replication.
We measure the effectiveness of an intervention by calculating how much it reduces the performance of a standard authorship attribution classifier. This reduction is measured relative to an estimate of how well the same classifier would have performed had the defensive technique not been used.

\section{Reproduction\label{sec:reproduction}}
In this section, we report on a reproduction of the study conducted by \citet{brennan2012adversarial} using its methods and corpus, the Extended Brennan-Greenstadt corpus.

\subsection{Extended Brennan-Greenstadt Corpus\label{sec:ebg}}

The Extended Brennan-Greenstadt corpus (EBG) contains writing from 45 individuals recruited on MTurk no later than 2012.
Two distinct types of writing were collected from participants: pre-existing samples and responses to writing prompts.

For the pre-existing samples, each participant uploaded at least 6,500 words of formal writing.
The participants were instructed not to upload writing containing extensive ``dialog/quotations'' or samples ``less than 500 words, laboratory and
other overly scientific reports, Q\&A-style samples such as exams, [or] anything
written in another person's style'' \citep{brennan2012adversarial}.

After uploading writing samples, each participant was asked to write two short essays responding to two different writing prompts. Accompanying each prompt were instructions on how to modify one's writing style in order to frustrate someone trying to identify the essay's author on the basis of their writing style. The first prompt asked the participant to write ca. 500 words describing their neighborhood to someone who has never been there. The prompt is prefaced with the instruction that the participant should try to hide their identity by changing their style of writing.
No suggestions are given to the participant regarding how they should go about the change. This strategy is labeled the obfuscation strategy.

For the second prompt, participants were asked to describe a day in their life using third-person narration. To conceal their writing style, they were instructed to imitate the distinctive writing style of the novelist Cormac McCarthy. This is labeled the imitation strategy. Participants were given a 2,500-word excerpt of McCarthy's writing from \textit{The Road} and told to read the excerpt before composing their response.

Table~\ref{tbl:corpus} summarizes the Extended Brennan-Greenstadt corpus. Additional details can be found in \citet{brennan2012adversarial}.

\begin{table}[!ht]
\caption{Summary of corpora used in the reproduction and replication experiments. Text lengths are reported in words.\label{tbl:corpus}}
 \small
  \centering
  \begin{tabular}{ccccc}
    \toprule
    Corpus     & Task     & Authors     & Avg. Training Length     & Avg. Testing Length \\
    \midrule
    
\multirow{2}{*}{EBG}    & Obfuscation                   &   45         & 8,727          & 564   \\
                        & Imitation                     &   45         & 8,727          & 574   \\
\cmidrule(r){1-5}
\multirow{3}{*}{RJ}     
& Control                       &   21         & 7,064         & 582   \\
& Obfuscation                   &   27         & 7,829         & 570   \\
                        & Imitation                     &   17         & 7,752         & 583   \\
    \bottomrule
  \end{tabular}
\end{table}

\subsection{Method}

\citet{brennan2012adversarial} examine how authorship attribution classifier accuracy declines when a user attempts to conceal their writing style by using a defensive technique. Three different authorship attribution defenses are considered.
For each defense, \citet{brennan2012adversarial} examine how the size of the candidate pool affects classifier accuracy by sampling 1,000 sets of candidate authors from the pool of 45 authors. 
They consider sets of size 5, 10, 15, 20, 25, 30, 35, and 40, each drawn randomly from the candidate pool without replacement.
For each set, authorship attribution models are trained using the candidates' pre-existing writing samples as training data. The models are then asked to predict the author of essays composed in response to the prompts. In other words, the elicited essays form the test sets. The performance of the classifier on the test set is compared with a baseline: classifier accuracy on the pre-existing writing samples, where performance is measured using 10-fold cross-validation.
Our reproduction followed the exact same setup.

The effectiveness of the round-trip translation strategy was evaluated briefly in \citet{brennan2012adversarial} with a completely different corpus (the Brennan-Greenstadt corpus).
The translated texts and implementation details cannot be recovered.
Therefore, we did not examine the round-trip translation strategy in the reproduction study; we revisit this strategy in the replication study below.
We also take the liberty of reporting results obtained using two additional authorship attribution models, one using a simpler feature set and another using a recently-developed neural network classifier.

\subsubsection{Writeprints-static and Support Vector Machine Classifier}

\citet{brennan2012adversarial} measured the effectiveness of authorship attribution using three models.
A support vector machine (SVM) model with a polynomial kernel using the ``Writeprints-static'' feature set proved by far the most successful. This was not unexpected: the other models considered were unorthodox and have seen limited use by other researchers. Because the SVM model has been widely used and because it proved most successful in the study, we use this model in the reproduction study.

\paragraph{Writeprints-static feature set}

The Writeprints-static feature set is a simplified version of the ``Writeprints'' feature set proposed by \citet{abbasi2008writeprints}. The feature set includes 557 fixed (``static'') lexical and syntactic features; among these are frequent character bi- and trigrams, part-of-speech tags, and 403 function words.

We carefully re-implemented the Writeprints-static feature set in Python by consulting Jstylo's GitHub repository \citep{mcdonald2012use}.\footnote{See \url{https://github.com/psal/jstylo}.} Our re-implementation uses 552 features.
Features in the original study were recovered precisely with minor exceptions. Where a feature could not be recovered exactly, a close substitute was used.\footnote{These differences come in two groups. First, the most frequent character bigram and trigram lists could not be recovered. We used the most frequent character bigrams and trigrams in the Brown corpus. Second, \citet{brennan2012adversarial} used a part-of-speech tagset consisting of 22 tags, which we cannot locate. We used the widely-used ``universal'' POS tagset V2, which consists of 17 tags. The Python package ``writeprints-static'' built for extracting the feature set is released on PyPI \url{https://pypi.org/project/writeprints-static}.}

\paragraph{Polynomial SVM}

\citet{brennan2012adversarial} report using an SVM model with a polynomial kernel but do not indicate the parameters they used.
To find suitable parameters, we performed a grid search on the training examples of the Extended Brennan-Greenstadt corpus (10-fold cross-validation).
The range of parameters searched follows the suggestions of the LIBSVM authors \citep{CC01a}.
 
We chose an SVM model using a polynomial kernel with the following parameters: degree ($d$) equal to 3, regularization (``C'') equal to $0.01$, $\gamma$ equal to $0.001$, and constant bias $r$ equal to 100.
Many sets of parameters performed roughly as well as these.
The chosen polynomial SVM has an average accuracy of 83.0\% in attributing authorship given 45 candidates (10-fold cross-validation)---virtually identical to the accuracy reported in the original study.

Most of the Writeprints-static features are counts (e.g., of POS tags or function words). Because documents differ in length, we normalize each document's feature vector by the sum of its elements. Each feature is then standardized by dividing by the standard deviation after mean-centering.

\subsubsection{Koppel-512 with Logistic Regression}

The ``Koppel-512'' function word list contains 512 function words adopted from the widely-cited authorship attribution experiment described in \citet{koppel2009computational}.
Function words are typically free of obvious meaning (e.g., ``the'', ``and'', ``or'', and ``this'') and have been used extensively in authorship attribution research \citep{kestemont2014function}.
We use standard multi-class logistic regression with quadratic regularization ($\lambda = 1.0$).
Function word frequencies are normalized and standardized using the same preprocessing as for the Writeprints-static features. The classifier performance is roughly similar to that of the Writeprints-static-based SVM model.

We include this model chiefly because we anticipate that future researchers interested in reproducing our results will have no difficulty extracting features from the texts in a way that matches our implementation. In particular, extracting features using the Koppel-512 feature set should be considerably easier than extracting features using the Writeprints-static feature set.

\subsubsection{RoBERTa}

We further adopt a RoBERTa model \citep{liu2019roberta} pretrained with five large corpora (the Book Corpus, English Wikipedia, Common Crawl-News, OpenWebText, and Stories corpora).
The model (``roberta-base'') is provided by \citet{wolf2020transformers}.
We adapt the model for classification by continuing to train using our data, updating only the parameters in a final layer for classification.
We use a low learning rate ($3\text{e--}5$), and all samples are padded or truncated to 512 tokens, as the model requires.
During training, we hold out the first training example from each of the candidates to create a validation set.
The model is trained until validation loss fails to improve for 50 epochs.
In general, this occurs after no more than 200 epochs.
We use parameters associated with the lowest validation loss.

Note that we use this model slightly differently than we do the other models.
First, with RoBERTa, we refrain from running cross-validation on the training data because cross-validation is not a standard practice when using deep learning models.
Second, to reduce computational cost, we consider only ten (instead of 1,000) runs at each candidate size.

\subsection{Results}

\begin{figure}[!ht]
\includegraphics[scale=0.85]{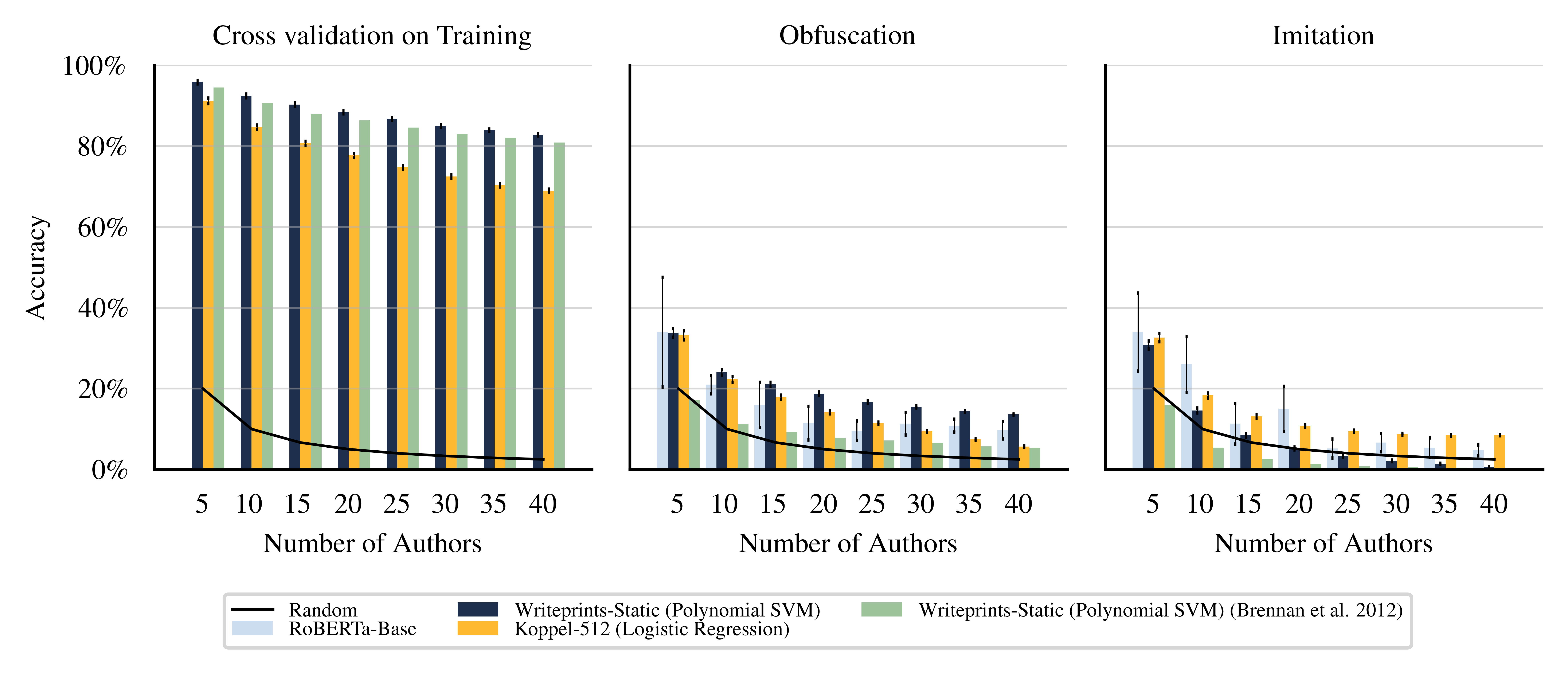}
\caption{Reproduction of the adversarial stylometry experiment in \citet{brennan2012adversarial}. The left panel shows the accuracy of a 10-fold cross-validation using the training data.
The middle and right panels indicate the accuracy of the classifier when the indicated authorship attribution circumvention strategy is used.
Each bar indicates the mean accuracy; error bars show 95\% confidence intervals. 10-fold cross-validation is not performed with the RoBERTa model.\label{fig:reproduction}}
\end{figure}

The results of the reproduction study are summarized in Figure~\ref{fig:reproduction}, whose three panels correspond to Figures 6--8 in \citet{brennan2012adversarial}.
For comparison with the original, we show the accuracy statistics reported in the original paper side-by-side with our results.\footnote{We recovered the accuracy statistics directly from their figures using WebPlotDigitizer \citep{Rohatgi2020}.}

The left panel of Figure~\ref{fig:reproduction} shows classifier accuracy on the training data measured using 10-fold cross-validation. 
The middle and right panels show the accuracy of the authorship attribution classifier when the subject uses the indicated defensive strategy.
As we saw in the original paper, classifier performance drops dramatically when either defensive strategy is used, and the imitation strategy is more successful than the obfuscation strategy at confusing the Writeprints-static-based SVM classifier.

\section{Replication\label{sec:replication}}

To further increase confidence in the original result, we replicate the study in \citet{brennan2012adversarial}, re-running the experiment with new participants.\footnote{Our experiment was approved by the Institutional Review Board of Indiana University (No. 1806940835).}
We label the new corpus of writing samples, analogous to the Extended Brennan-Greenstadt corpus, the Riddell-Juola corpus. In our replication, we correct two conspicuous flaws in the original experiment design: the lack of a control group and the non-random assignment of writing prompts. We resolve the first flaw by introducing a control condition and the second by using a single writing prompt, so each participant produced one essay rather than two essays from each participant.

In addition to the formal replication, we informally explore the round-trip translation defense discussed in the original study.

\subsection{Riddell-Juola Corpus}

The Riddell-Juola Corpus (RJ) was gathered using essentially the same procedure as that used to compile the Extended Brennan-Greenstadt corpus \citep{brennan2012adversarial}. The obfuscation and imitation strategy instructions from the original study were reused, as was the describe-your-neighborhood writing prompt.\footnote{
Instructions and prompts used in the original study remain available at the original URL: \url{https://psal.cs.drexel.edu/tissec/CorpusParticipation.txt}.}

As in the original study, participants were recruited on MTurk.
About 6,500 words of pre-existing formal writing were solicited with the same instructions used in the original study. These writing samples comprise the training data for the authorship attribution classifier.

For the ca. 500-word essay, participants were instructed to respond to the describe-your-neighborhood writing prompt. The beginning of each prompt reads: ``You are asked as part of a college application to describe your neighborhood to someone who has never been there before.''

In contrast to the original study, participants were randomly assigned to receive no additional instruction (\textit{control}), the obfuscation strategy instruction, or the imitation strategy instruction. Each participant only submitted a single 500-word essay in response to the describe-your-neighborhood prompt. This design eliminates concerns about order effects and the relative ease (or difficulty) of ``executing'' specific strategies with particular prompts.\footnote{A disadvantage of this new design is that each participant contributes only one essay. In retrospect, we realize there is a more cost-effective design available: ask respondents to write more than one essay but randomly assign the writing prompt as well as the defensive strategy.}

As in the original study, the authorship attribution model must predict the authorship of these ca. 500-word essays.

\subsubsection{Round-trip Translation}

In the replication study, we leverage the essays composed in the control condition to evaluate how well round-trip translation conceals participants' writing style. We use the Google Translate API to translate between languages.\footnote{The API is wrapped in the Python package ``translators'' (v.4.9.5).}

We use the intermediate language choices from \citet{brennan2012adversarial}: the two single-step translations are English--German--English and English--Japanese--English; and the two-step translation is English--German--Japanese--English.
Note that, when the original study was performed, Google was using statistical machine translation. The current system, introduced in 2016, uses neural machine translation.

\subsubsection{Demographic Characteristics}

Responses were collected between March 29th and June 1st, 2019.
Respondents completed a demographic questionnaire, reporting their gender and age bracket (Table~\ref{tbl:demographics}).
We excluded responses that were not in English or that appeared inauthentic.
The pre-existing writing samples were further processed in order to remove personally identifying information.
Lengthy quotations, headings, tables, and figures were also removed.
See Table~\ref{tbl:corpus} for statistics describing the corpus. 
Participants provided informed consent by agreeing to participate after reading the study description on MTurk, which stated the purpose of the research and the nature of the tasks.

\begin{table}[!ht]
 \caption{Demographic characteristics of participants contributing writing samples and essays to the Riddell-Juola corpus. Participants were assigned an authorship attribution circumvention strategy (or to the control group) uniformly at random.\label{tbl:demographics}}
 \small
  \centering
  \begin{tabular}{ccccc}
    \toprule
    Demographics     & Attribute     & Obfuscation  & Imitation   & Control \\
    \midrule
\multirow{2}{*}{Gender} & Woman             &   13         & 10          & 10         \\
                        & Man               &   14         & 7           & 11         \\
\cmidrule(r){1-5}
\multirow{3}{*}{Age}    & 18--34           &   18         & 12          & 16         \\
                        & 35--49           &   7          & 5           & 3          \\
                        & 50--64           &   2          & 0           & 2          \\
    \bottomrule
  \end{tabular}
\end{table}

\subsection{Method}

In the replication, we adopted the same feature set, model, and setup as used in the reproduction.

\subsection{Results}

\begin{figure}[!ht]
\includegraphics[scale=0.86]{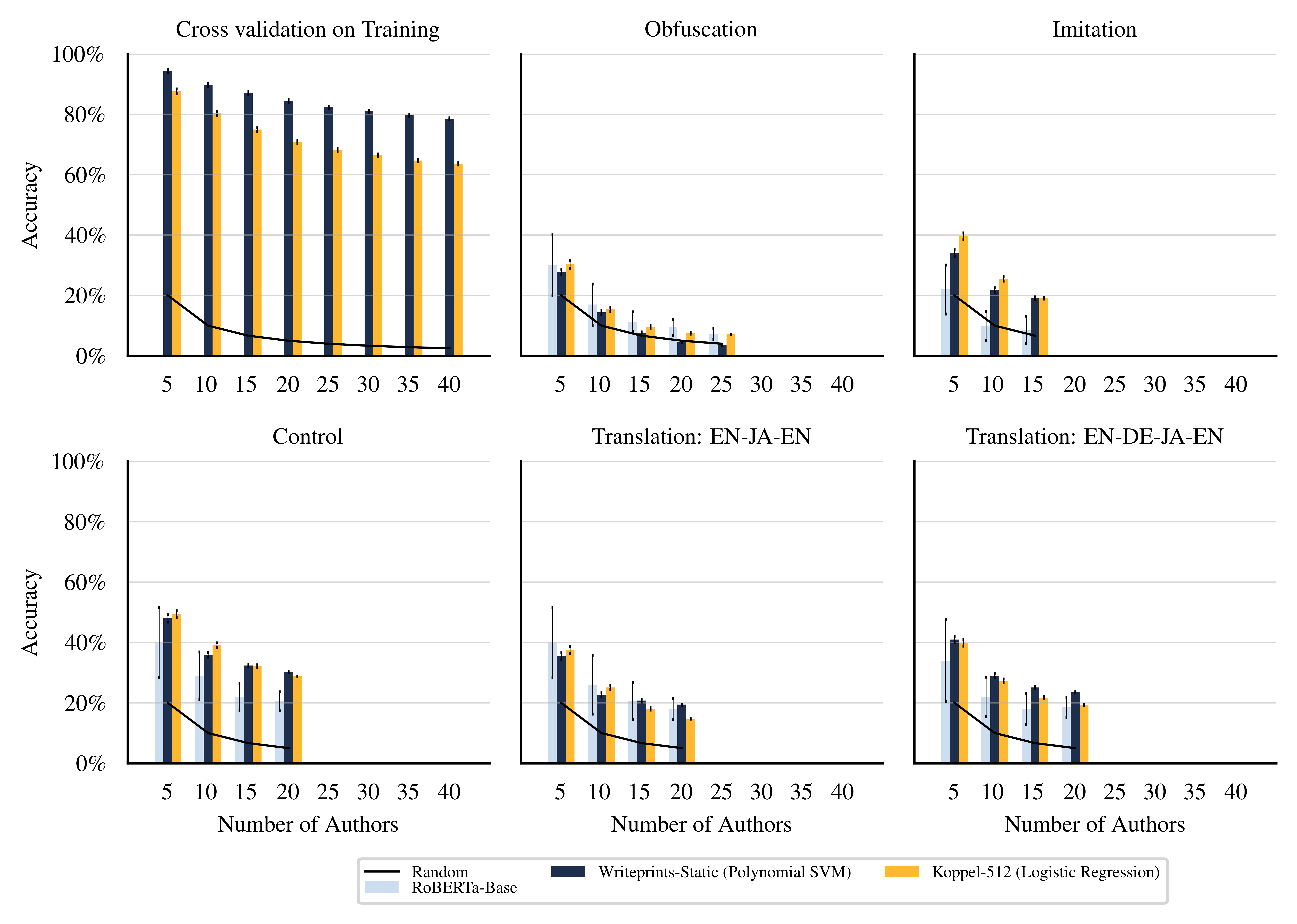}
\caption{Replication of the adversarial stylometry experiment in \citet{brennan2012adversarial} using the Riddell-Juola corpus. The top panels show classifier accuracy on the training data measured using 10-fold cross-validation, the obfuscation strategy, and the imitation strategy.
The bottom-left panel shows classifier accuracy on writing from the control group.
The bottom-middle and bottom-right panels show the accuracy of two round-trip translation strategies.
Each bar indicates the mean accuracy with candidate sets randomly sampled 1,000 times (without replacement) from the pool of participants who used the indicated strategy.\label{fig:replication}}
\end{figure}

The replication study generally confirms the effectiveness of the circumvention techniques reported in the original paper.
The results of the replication experiment using the new Riddell-Juola corpus are summarized in Figure~\ref{fig:replication}.

The upper-left panel of Figure~\ref{fig:replication} shows the accuracy of 10-fold cross-validation on the training data.
Model performance slowly decreases as candidate count increases.
However, with cross-validation, authorship attribution models can leverage topical information in documents in making predictions. (Submitted pre-existing writing samples often concern similar subjects.)
Hence, the accuracy measured using the control group (bottom-left panel) is a more appropriate baseline.
Recall that participants in the control group responded to a fixed topic, and no suggestion was made that they should modify their style.
Because every participant in the control group writes on the same topic, the risk that topics overlap with those in pre-existing writing samples is negligible.

The upper-middle and upper-right panels indicate the performance of the models when the corresponding strategy is used. Due to random assignment in the Riddell-Juola corpus, the number of writers using each strategy varies. When an authorship attribution circumvention strategy is used, classifier accuracy suffers relative to the control group.
In contrast to the results obtained using the EBG corpus, the obfuscation strategy appears to be more effective than the imitation strategy.

The results concerning the round-trip translation strategies are shown in the bottom-middle and bottom-right panels of Figure~\ref{fig:replication}.
We omit plotting German as the intermediate language because the strategy performs as well as the strategy using two intermediate languages. The round-trip through Japanese performs best among the round-trip translation strategies.
The accuracy reduction achieved by the round-trip translation strategy is roughly the same as that achieved by the imitation strategy.

\subsection{Examination of Translated Samples}

We examined the translated samples to make sure that their meaning remained consistent with the originals.
In general, the meaning of sentences was preserved; poor translations were often attributable to unconventional language use and misspellings in the original prose, and drastic changes in meaning were rare. (See Table~\ref{tbl:translations} for examples of typical and infelicitous translations.)
When using one intermediate language, semantics tend to be more faithfully preserved.
Perfect reproduction of the original text is occasionally observed, especially for short sentences (e.g., Sample 5 in Table~\ref{tbl:translations}).
We found that using Japanese as the intermediate language tended to produce fewer perfect round-trip translations, something desirable in the present context.

Misspellings hurt the performance of the round-trip translation defense: they may be copied verbatim into the translated samples (Table~\ref{tbl:translations}, Sample 11), which could reveal an individual's identity if they tend to misspell particular words.
Misspellings also appear to contribute to semantic loss in translation (Sample 10) and lead to other mistranslations (Samples 12 and 13).

In short, this strategy appears to work well when original sentences are relatively simple and grammatical.

\begin{table}
\caption{Observations from the translation samples. Obvious changes are marked in italic.\label{tbl:translations}}
 \small
  \centering
  \begin{tabular}{p{2cm}p{0.3cm}p{4cm}p{1.5cm}p{4cm}}
    \toprule
    Observations     & No.      & Original       & Route       & Translation  \\
    \midrule
    Synonym \mbox{substitution}         
                                 & 1
                                 & It's set upon a hill, surrounded by buildings with historical meaning.
                                 & EN-DE-EN
                                 & It \textit{is on} a hill, surrounded by buildings with historical \textit{importance}. \\
    \cmidrule(r){2-5}            
     Paraphrasing                & 2
                                 & We were just young kids without a care in the world.        
                                 & EN-JA-EN    
                                 & We were young children without \textit{worrying about} the world. \\
                                 & 3
                                 & Since I live in a large city, my neighborhood is extremely diverse. 
                                 & EN-DE-JA-EN
                                 & \textit{I live in a big city, so my neighborhood is very diverse.} \\
    \cmidrule(r){2-5}
    Simplification               & 4
                                 & However, nowadays, a lot of people just use apps just as GrubHub and DoorDash to get their food delivered to them. 
                                 & EN-JA-EN    
                                 & \textit{But today, many people} use apps \textit{like} Grubhub and Doordash to \textit{deliver food}. \\
    \cmidrule(r){1-5}
``Perfect'' back translation     & 5
                                 & It is a really perfect place to live.        
                                 & EN-JA-EN    
                                 & It is a really perfect place to live.  \\
    \cmidrule(r){2-5}
    Some semantic loss           & 6
                                 & There are countless cafés that were home to many hours of caffeine fueled cramming.
                                 & EN-DE-EN    
                                 & There are countless cafés \textit{in which many hours of caffeine were founded}.  \\
                                 & 7
                                 & The lawns are kept in decent condition. Nobody has leaves piled up from fall time.
                                 & EN-JA-EN
                                 & The lawn is kept in a decent state. \textit{Nobody has piled up since the fall time}. \\
    \cmidrule(r){2-5}
    Unacceptable semantic loss   & 8
                                 & She was from a town in which the grocery store was a 5 minute walk from her house... 
                                 & EN-DE-JA-EN 
                                 & She came from her house 5 minutes from her house...      \\
                                 & 9
                                 & ...but I ended up getting close to someone who lives two houses down from me and another guy that lives on my street.
                                 & EN-JA-EN
                                 & ...but I \textit{approached me from another man who lived} on my street. \\
                                 & 10
                                 & ...i am a drywall and T-bar instalator...
                                 & EN-DE-EN 
                                 & ...I am a drywall and T-bar \textit{information}...\\
    \cmidrule(r){2-5}
    Misspelling copying          & 11
                                 & The people don't feel as optomistic as they used to... 
                                 & EN-DE-EN 
                                 & People don't feel as optomistic as before...      \\
    \cmidrule(r){2-5}
    Downstream mistranslation         & 12
                                 & The streets are wide and the kids usually play footbal... 
                                 & EN-DE-EN 
                                 & The streets are wide and the children usually play \textit{foot bales}...      \\
                                 & 13
                                 & The streets are wide and the kids usually play footbal... 
                                 & EN-DE-JA-EN 
                                 & The street is large, and the children usually play \textit{the BA on the feet}...      \\
    \bottomrule
  \end{tabular}
\end{table}

\section{Discussion}

Overall, we consider this a successful replication of the primary claim: both manual strategies substantially reduced authorship attribution accuracy in our independent sample, consistent with the original finding. 
One discrepancy is that the relative performance of the two strategies is reversed; obfuscation outperforms imitation in our data, contrary to the original study.

We speculate that this may be due to one or both of the following factors.
First, the manual intervention writing samples vary greatly in quality due to the diversity of participants on MTurk.
Individuals may simply vary in their ability to use the defense effectively.
Second, the prompts used for eliciting writing samples using the imitation defense are different.
It may be easier in some sense to use the defense when writing about one's day from a third-person perspective than when writing about one's neighborhood.

Further study of the round-trip translation strategy---or some similar language-model-based technique such as GPT-2 \citep{radford2019language} or T5 \citep{raffel2020exploring}---is warranted for three reasons. First, the technique merits attention because it requires no human intervention. This means it can be used in settings where the writer is unavailable or lacks time to perform a manual defense.
Second, existing flaws such as mishandling of misspellings seem correctable given advances in language modeling \citep{karpukhin2019training}.
Third, because the defense is automatic, it may be combined with a manual intervention---delivering, potentially, an incrementally more potent defense.

If it proves effective, the round-trip translation must be usable offline. Relying on an online API, as we do here, would be an unacceptable risk for a whistleblower attempting to conceal their identity from a government or multinational firm that closely monitors IP traffic.

\section{Conclusion}

This study investigated the effectiveness of three adversarial stylometry strategies: obfuscation, imitation, and round-trip translation.
We estimated how much each strategy reduces the performance of a standard authorship attribution model. This study generally confirms the findings of \citet{brennan2012adversarial}: asking an individual to try to conceal their writing style yields prose that is more difficult for an authorship attribution model to link with the writer's preexisting writing.
For example, in the setting where a classifier must predict the author of a text given ten candidates, performing either manual strategy reduces classifier accuracy from $\approx$40\% to $\approx$20\%. This is a meaningful reduction. If these results generalize, an adversary using standard techniques to identify an individual who has used one of the manual defenses will learn less about the likely author of an unsigned text than they otherwise would. An adversary committed to acting based on the classifier's prediction will have a higher risk of incorrectly identifying a writer who is not the author of the unsigned text.

\section*{Social Impact \& Responsibility}
This work studies defenses against authorship attribution, a procedure that can result in unwanted deanonymization of writers. Understanding when common defenses do and do not work can benefit individuals who have real needs for anonymous communication (e.g., whistleblowers and journalists). These defenses are dual use, but we expect their net effect to strengthen privacy for writers at risk; we highlight limitations and practical safety guidance.

Our replication highlights a concrete safety consideration: using third-party online translation services may introduce additional privacy risks (e.g., metadata leakage or traffic monitoring). For high-risk users, any automated defense should be usable offline, and users should avoid relying on online APIs in adversarial settings.

\section*{Transparency Statement}
We report how we determined our sample size, all data exclusions (if any), all manipulations, and all measures in the study.

Sample size and inclusion/exclusion: For the reproduction, we analyzed the full Extended Brennan-Greenstadt (EBG) corpus (45 authors). For the replication, we recruited MTurk participants between March 29 and June 1, 2019 and analyzed all submissions meeting inclusion criteria; non-English, non-responsive, and likely inauthentic responses were excluded. Pre-existing writing samples were processed to remove personally identifying information. Participant-level metadata (self-reported gender, age bracket, task duration, collection date, and exclusion notes) are provided in \texttt{metadata.csv}.

Manipulations and measures: In the replication, participants were randomly assigned to the control condition or to one of two manual defense instructions (obfuscation or imitation). We additionally evaluated round-trip translation on control essays using the Google Translate API. The primary outcome is authorship attribution accuracy under the evaluation protocol described in the manuscript.

Connection to the original work: The authors have no overlapping authorship with the original study and no formal collaboration with the original authors on this manuscript.

Competing interests: The authors declare no competing interests.

Use of AI: Claude 4.6 (Opus) was used to proofread sections after the Conclusion (Social Impact \& Responsibility, Transparency Statement, etc.); all analyses, claims, and final text were reviewed and verified by the authors.

File-drawer statement: The authors report all reproduction and replication studies and analyses carried out for this manuscript.

\section*{Acknowledgments}

This material is based upon work supported by the National Science Foundation under Grant No. 1814425. Any opinions, findings, and conclusions or recommendations expressed in this material are those of the author(s) and do not necessarily reflect the views of the National Science Foundation.

\section*{Author Contributions}
Haining Wang: Data Curation; Investigation; Software; Formal analysis; Visualization; Writing -- Original Draft; Writing – Review \& Editing.
Patrick Juola: Funding Acquisition; Writing -- Review \& Editing.
Allen Riddell: Conceptualization; Data Curation; Funding Acquisition; Methodology; Project Administration; Resources; Validation; Supervision; Writing – Original Draft; Writing -- Review \& Editing.

\section*{Preregistration}
None of the reported studies were preregistered. 

\section*{Data, Materials, and Code Availability}
A frozen reproducibility bundle (code, scripts, replication data, and study materials) is archived on Zenodo: \url{https://doi.org/10.5281/zenodo.18729526}. The archive contains \texttt{rr\_bundle.zip} and records exact code/data commits in \texttt{FROZEN\_COMMITS.txt}. Replication participant metadata are provided in \texttt{metadata.csv} within the archive. The MTurk study materials (prompt text and survey instrument) are included in the same archive.

We do not redistribute EBG in our Zenodo archive as it is a third-party dataset distributed by the original authors. EBG is available from the Anonymouth repository at \url{https://github.com/psal/anonymouth/tree/master/jsan_resources/corpora/amt}. Our code expects EBG to be placed under \texttt{resource/Drexel-AMT-Corpus/} (see the Zenodo README for details).

Licenses: Code in the Zenodo archive is released under the ISC license (see \texttt{LICENSE}). The Riddell-Juola corpus and associated metadata and study materials are released under CC0 1.0 (see \texttt{DATA\_LICENSE}).

\bibliography{references}  
\bibliographystyle{acl_natbib}

\end{document}